\documentclass[10pt, a4paper]{article}

\usepackage[final]{lrec2026} 

\usepackage{hyperref}
\usepackage{url}
\usepackage{booktabs}
\usepackage{multirow}
\usepackage{graphicx}
\usepackage{xcolor} 
\usepackage{enumitem}

\setlist[itemize]{noitemsep, topsep=0.5pt}

\newcommand{\dataname}{\textsc{InData}}

\title{InData: Towards Secure Multi-Step, Tool-Based Data Analysis}

\name{Karthikeyan K, Raghuveer Thirukovalluru, Bhuwan Dhingra, David Edwin Carlson} 

\address{Duke University \\
         \texttt{\{karthikeyan.k,raghuveer.thirukovalluru,bhuwan.dhingra,david.carlson\}@duke.edu}\\}

\abstract{
Large language model agents for data analysis typically generate and execute code directly on databases. However, when applied to sensitive data, this approach poses significant security risks. To address this issue, we propose a security-motivated alternative: restrict LLMs from direct code generation and data access, and require them to interact with data exclusively through a predefined set of secure, verified tools. Although recent tool-use benchmarks exist, they primarily target tool selection and simple execution rather than the compositional, multi-step reasoning needed for complex data analysis. To reduce this gap, we introduce Indirect Data Engagement (\dataname), a dataset designed to assess LLMs’ multi-step tool-based reasoning ability. \dataname\ includes data analysis questions at three difficulty levels—Easy, Medium, and Hard—capturing increasing reasoning complexity. We benchmark 15 open-source LLMs on \dataname\ and find that while large models (e.g., gpt-oss-120b) achieve high accuracy on Easy tasks (97.3\%), performance drops sharply on Hard tasks (69.6\%). These results show that current LLMs still lack robust multi-step tool-based reasoning ability. With \dataname, we take a step toward enabling the development and evaluation of LLMs with stronger multi-step tool-use capabilities. We will publicly release the dataset and code.\newline ,Tools, Systems, Applications }

\begin{document}

\maketitleabstract

\section{Introduction}

Large Language Models (LLMs) have demonstrated strong capabilities in program synthesis \citep{jiang2024survey, le2022coderl}, mathematical reasoning \citep{ahn-etal-2024-large, pan-etal-2023-logic}, and data analysis \citep{hong2024datainterpreterllmagent, yang2024matplotagent}. These capabilities drove rapid adoption across industries that handle sensitive information, including healthcare and finance \citep{cascella2023evaluating, li2023large}, where organizations increasingly rely on LLMs to assist with tasks such as querying databases, generating insights.

However, most commercial LLMs operate as external services that require transmitting data to their servers for processing. When analysts send sensitive datasets—such as patient medical records or financial transactions—to external LLM providers, they lose control over how the data is stored, processed, and retained \citep{yao2024survey, das2025security}, potentially violating data use agreements and regulatory requirements like HIPAA or GDPR. To address these privacy concerns, existing work on LLM agents for data analysis predominantly adopts a code-generation approach: the LLM receives only a description of the analysis task and generates executable code (typically Python or SQL) that processes the data locally, ensuring sensitive data never leaves the protected environment \citep{Lai2022DS1000, hong2024datainterpreterllmagent}.

While code-generation mitigates privacy risks, it introduces severe execution risks: (1) the generated code may invoke external APIs that leak sensitive data; (2) it may contain errors that corrupt data \citep{zhang-etal-2024-benchmarking-data}; (3) it could introduce security vulnerabilities\citep{basic2025largelanguagemodelscode}; and (4) it often lacks proper error handling and logging, making failures difficult to audit or debug. Motivated by these risks, we propose restricting LLMs from generating or executing arbitrary code, instead constraining them to solve user task using only a predefined set of vetted tools that securely handle data locally. This approach addresses both privacy and execution risks: since the LLM never receives the raw data and cannot generate arbitrary code, sensitive information remains protected while eliminating execution risks from LLM generated code. While recent work has explored LLM tool-calling capabilities \citep{qin2024toolllm}, existing benchmarks primarily evaluate tool selection from large API collections or simple tasks requiring few tool calls. No existing benchmark evaluates LLMs’ ability to solve complex, multi-step tool-calling tasks. 

\begin{figure}
    \centering
    \includegraphics[width=0.95\linewidth]{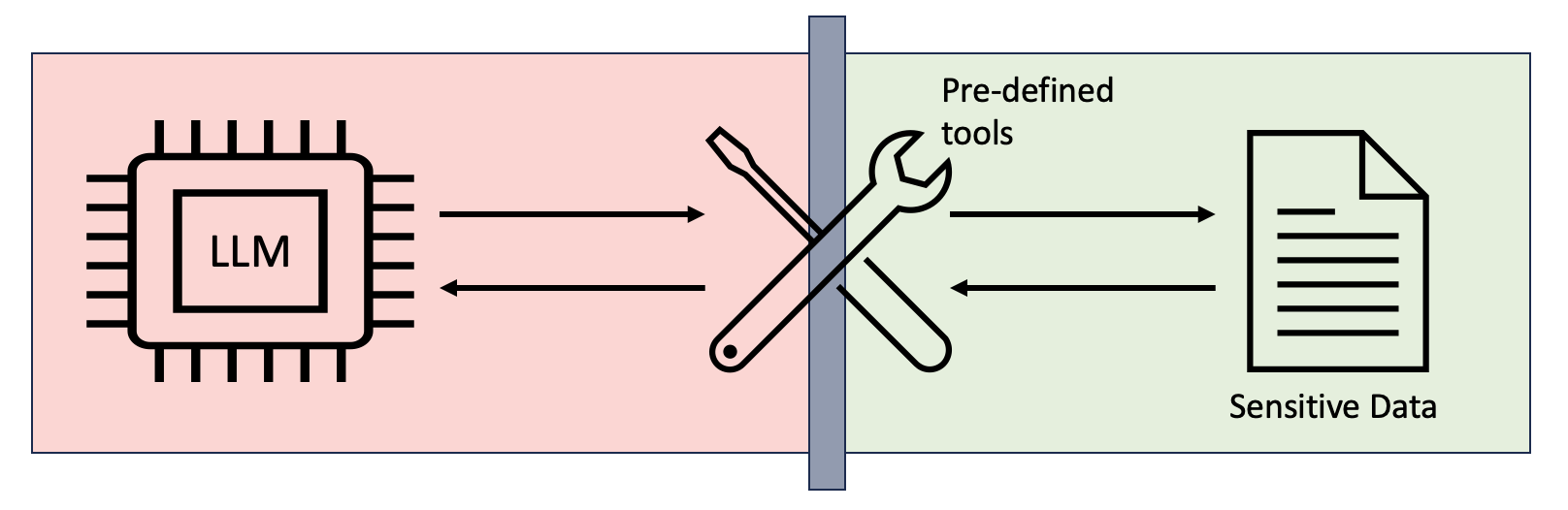}
    \caption{Predefined tools act as a secure barrier between the LLM and sensitive data.}
    \label{fig:intro}
\end{figure}

To address this gap, we introduce \dataname, designed to evaluate LLMs' ability to solve complex, multi-step reasoning data analysis problems using only a predefined set of tools. Our dataset consists of four components: (1) multi-step reasoning questions that require sequential tool usage to solve, (2) underlying CSV data on which the questions rely, (3) a curated set of tools for data manipulation and analysis, and (4) ground truth answers for evaluation. \dataname\ comprises 100 tools, 114 CSV files, and 2063 questions spanning three difficulty levels: 484 easy, 369 medium, 1210 hard. Typically easy questions require fewer than 5 tool calls, whereas hard questions often demand more than 15 sequential tool calls to reach the final answer.

We benchmarked 15 open-source LLMs on \dataname. Our results show that while LLMs handle coding and simple tool use—areas well covered by existing benchmarks—they struggle with compositional, multi-step tool reasoning. We also observe a stark performance gap across model sizes. Smaller models, such as Qwen3-1.7B, achieve only 7.5\% accuracy on hard questions, whereas gpt-oss-120b, reaches 69.6\%. Even among state-of-the-art models, performance declines sharply with problem difficulty—gpt-oss-120b drops from 97.3\% on easy questions to 69.6\% on hard ones. 

Furthermore, while tool-only approaches enable secure data analysis, current LLMs lack the capability for complex, multi-step tool-based reasoning, partly because no existing benchmarks systematically evaluate this ability, and consequently, models are not optimized for it. \dataname\ addresses this gap by providing a benchmark that measures and facilitates progress in LLMs’ multi-step tool reasoning capabilities. Additionally, we perform extensive analyses and find that: (1) structured hints, such as Python code solutions, substantially improve multi-step tool reasoning; (2) larger models perform best when provided with all the available tools, whereas smaller models perform best with only a limited, relevant subset of tools; and (3) model families differ in behavior—GPT-based models tend to retry persistently, while Qwen-based models stop earlier.

Our contributions include: (1) we propose a security-first, tool-only methodology for LLM-based data analysis; (2) we introduce \dataname, the first dataset specifically designed to evaluate LLMs' multi-step, tool-based reasoning ability; (3) we benchmark 15 open-source LLMs on \dataname\ and observe that while current models are proficient in code generation and simple tool usage, they lack compositional, multi-step tool reasoning ability; and (4) we conduct an extensive analysis and outline directions for improving LLMs’ tool reasoning ability.

\section{Related Works}


\paragraph{LLMs for Data Analysis: Datasets}Advances in the agentic capabilities of large language models (LLMs) have led to new datasets and frameworks for evaluating their performance on data analysis tasks. Early contributions include DS-1000 \citep{Lai2022DS1000}, a benchmark of 1,000 Python data science problems from Stack Overflow, and ARCADE \citep{yin-etal-2023-natural}, which features 1,078 Jupyter notebook-based tasks. Both primarily assess code-generation proficiency. More recent efforts, such as DAEval \citep{pmlr-v235-hu24s}, which evaluates 34 LLMs across 257 closed-form analysis questions, and TAPILOT-CROSSING \citep{li2025benchmarking}, which focuses on conversational tabular data analysis, emphasize agentic reasoning through iterative code generation and refinement in a sandbox. Complementary benchmarks in program synthesis \citep{austin2021programsynthesislargelanguage, chen2021codex, hendrycksapps2021, DBLP:journals/corr/abs-2102-04664} and text-to-SQL conversion \citep{li2023llmservedatabaseinterface, Yu&al.18c, Yu&al.19, yu-etal-2019-cosql} similarly require models to produce executable code. In contrast, our work diverges by explicitly prohibiting code generation or execution, compelling LLMs to complete data analysis tasks solely through the use of predefined tools.
\paragraph{LLMs for Data Analysis: Frameworks and Evaluation}Several frameworks and evaluation methodologies have emerged to assess and improve LLM agents in data analysis. Data Interpreter \citep{hong2024datainterpreterllmagent} introduced an LLM agent framework for solving end-to-end data science problems, where models primarily generate code while optionally leveraging external tools. DS-Agent \citep{DS-Agent} employed a case-based reasoning approach to automate machine learning model construction and training across 30 data science tasks. Other specialized frameworks target subtasks such as data cleaning—most notably CleanAgent \citep{qi2024cleanagent}, built on the DataPrep.clean module \citep{dataprepeda2021}, which enables LLMs to produce code for user-specified formatting instructions. In contrast, our work explicitly prohibits code generation entirely. Complementary evaluation efforts further benchmark LLM agents. DataSciBench \citep{zhang2025datascibenchllmagentbenchmark} offers 222 data science prompts with fine-grained metrics evaluating 23 models, while DSEval \citep{zhang-etal-2024-benchmarking-data} assesses agents such as Chapyter, ChatDev \citep{chatdev, colearning, macnet, iagents}, CoML \citep{zhang2023mlcopilot}, and Jupyter-AI. Additionally, \cite{rasheed2024largelanguagemodelsserve} explored multi-agent LLM frameworks in qualitative analysis contexts. In contrast to these datasets, frameworks, and evaluation methodologies—which fundamentally rely on LLMs' code-generation capabilities—our work prohibits code generation entirely and requiring LLMs to solve the tasks using only predefined tools.

\begin{table*}[!htbp]
\centering

\begin{tabularx}{\textwidth}{@{} p{1.45cm} X @{}}\toprule
\textbf{Difficulty} & \textbf{Question} \\
\midrule
Easy & Given the file \texttt{cleaned\_global\_water\_consumption.csv}, what is the maximum ``Per Capita Water Use (Liters per Day)'' recorded for `Japan'? \\
\cmidrule(l){1-2} 
Hard & Given the file \texttt{cleaned\_global\_water\_consumption.csv}, calculate the ratio of the 90th percentile of ``Agricultural Water Use (\%)'' to the 10th percentile of ``Industrial Water Use (\%)'' for all data from the year `2018'. However, before this calculation, you must first create a cleaned dataset by removing any rows from the original file where the ``Groundwater Depletion Rate (\%)'' is more than 2 standard deviations above the overall mean depletion rate for all years and countries. Then, from this cleaned dataset, further filter to only include countries where the ``Rainfall Impact (Annual Precipitation in mm)'' in `2018' was below the `2018' median. \\
\bottomrule
\end{tabularx}

\vspace{2ex} 

\begin{tabularx}{\textwidth}{@{} >{\raggedright\arraybackslash}p{4.4 cm} X @{}}
\toprule
\textbf{Tool Signature} & \textbf{Description} \\
\midrule
\texttt{get\_columns(filename)}
& Reads a CSV dataset and returns a list of column names. Assumes the file is in CSV format and can be read by \texttt{pandas.read\_csv}. \\

\texttt{get\_mean(filename, column)}
& Calculates the mean of a specified numerical column from a CSV dataset using \texttt{pandas}. Assumes that the column contains numeric values. \\

\texttt{filter\_rows(filename, column, operator, value)}
& Filters rows in a CSV dataset based on a condition on a specified column using a comparison operator and saves the filtered data to a new randomly named CSV file. Assumes numeric comparisons when possible. \\
\bottomrule
\end{tabularx}
\caption{Examples questions (top) and sample tool descriptions (bottom) from the \dataname\ benchmark.} 
\label{tab:indata-example}
\end{table*}

\begin{table}[!tbp]
  \centering
  \small
  \setlength{\tabcolsep}{3pt}
  \begin{tabular}{lccc}
    \toprule
    & \textbf{Easy} & \textbf{Hard} & \textbf{vHard} \\
    \midrule
    Num.\ Q (before) & 587 & 569 & 2260 \\
    Num.\ Q (after)  & 484 & 369 & 1210 \\
    Words/Q ($\mu\!\pm\!\sigma$) & 15.5$\pm$3.5 & 28.4$\pm$6.7 & 62.5$\pm$14.9 \\
    \bottomrule
  \end{tabular}
  \caption{\textbf{\dataname\ Dataset Statistics}: Num.\ Q: Total number of questions generated initially (before) and after filtering (after). Words/Q: Mean and std of average number of words per question.}
  \label{tab:data_stats}
\end{table}

\paragraph{Security and Privacy Concerns in LLM-Based Systems}Several studies reveal that LLM-generated code often contains vulnerabilities such as insecure API usage, poor input validation, and susceptibility to attacks \citep{basic2025largelanguagemodelscode, pearce2022asleep}, with developers using AI assistants sometimes introducing additional flaws \citep{perry2023users}. Beyond code security, LLMs also pose privacy risks: they can inadvertently memorize and reveal sensitive information from training data \citep{carlini2021extracting}, and transmitting proprietary or regulated data to external LLM services raises compliance and confidentiality concerns \citep{yao2024survey}. Moreover, LLM-generated code can produce runtime errors or corrupt data. \citep{zhang-etal-2024-benchmarking-data}. These risks motivate our tool-only approach, which prevents arbitrary code execution and keeps data within secure, controlled environments. 

\paragraph{LLMs with Tool Calling}A parallel research direction explores LLMs’ ability to interact with external tools and APIs. Early efforts such as APIBench \citep{patil2023gorillalargelanguagemodel, berkeley-function-calling-leaderboard} introduced collections of HuggingFace, TorchHub, and TensorHub APIs. Subsequent benchmarks—including API Bank \citep{li2023apibankcomprehensivebenchmarktoolaugmented}, Tool Alpaca \citep{tang2023toolalpaca}, ToolBench \citep{xu2023toolmanipulationcapabilityopensource}, ToolLLM \citep{qin2024toolllm}, and ToolACE \citep{liu2025toolace}—further expanded this line of work. Although these datasets feature extensive tool repositories (e.g., ToolACE with 26,507 tools), their complexity primarily stems from tool selection rather than deep multi-step reasoning \citep{qin2024toolllm}. In contrast, our dataset includes a relatively smaller set of 100 tools but requires substantially deeper reasoning, often more than 25 sequential tool calls to solve a single question. Tool-calling research has also extended to question answering and conversational contexts, including ToolQA \citep{zhuang2023toolqadatasetllmquestion} and Tooltalk \citep{farn2023tooltalkevaluatingtoolusageconversational}. However, despite some multi-step examples, Tooltalk remains limited in scale (78 examples), employs teacher forcing, and exhibits biased evaluations \citep{k2025studyleveragingsearchselffeedback}.

\section{Dataset Creation}

In this section, we describe \dataname\ creation pipeline 

\subsection{Tabular Dataset Selection}\label{sec:tabular_selection}

We selected 114 tabular datasets from Kaggle that met the following criteria: (1) a permissive license (MIT, Apache 2.0, or CC0); (2) a single CSV file; (3) total size under 1 MB; and (4) no more than ten columns. Text classification datasets were excluded. For each dataset, we collected the CSV file and any available column descriptions.

\subsection{Tool Generation}

A core component of \dataname\ is a predefined set of tools that LLMs can use to interact with tabular data and answer data analysis questions. Each tool includes a description explaining what action it performs and how to call it, along with executable code that carries out the action. We used an advanced LLM (Gemini 2.5 Pro) to generate 100 tool descriptions. Each description specifies the tool’s name, functionality, input arguments with their types and explanations, and the expected output type. We instructed the LLM to generate tool descriptions that satisfy three main requirements. First, each tool must perform a simple, atomic operation analogous to an individual \texttt{pandas} function, rather than complex compositions. This design makes the tools more broadly applicable and allows us to evaluate the LLM’s ability to compose multiple tools for multi-step reasoning. Second, tools may only read the CSV data—they are not allowed to modify it. This restriction helps prevent accidental corruption or deletion. Third, tools must not directly return datasets. Instead, any output DataFrames or Series objects must be written to a local temporary file, and only the filenames should be returned. For example, a tool such as \texttt{filter\_rows} writes the filtered result to a randomly named temporary file and returns its filename, which the LLM can reference in later tool calls. These constraints add a layer of protection against data leakage, even though we recommend running the LLM locally. To facilitate generation, we provided the LLM with a few example tool names (e.g., \texttt{get\_mean(filename, column)}, \texttt{get\_max(filename, column)}, \texttt{get\_columns(filename)}) as seeds. Refer Table~\ref{tab:indata-example} for example tools name and description.

\paragraph{Tool Code Generation: } For each tool description, we instruct Gemini 2.5 Pro to generate Python code that implements the specified functionality. We explicitly instruct the LLM that all generated functions must handle type errors robustly. Each tool must attempt to automatically resolve type mismatches—for example, when a CSV file contains a column named \textit{1} (integer) but the LLM provides the argument as "1" (string), the function should recognize and correct the discrepancy. When automatic resolution fails or any other runtime error occurs, the tool should return an informative error message. Before passing this message back to the LLM, we truncate it to 500 characters (the first and last 250) to minimize the risk of data leakage. In addition to the data analysis tools, we add seven special-purpose utilities: four arithmetic operations (\texttt{Add}, \texttt{Subtract}, \texttt{Multiply}, \texttt{Divide}) that operate on two numeric inputs, two reporting tools (\texttt{Report\_number}, \texttt{Report\_string}) that return final answers and terminate agent execution, and one control tool (\texttt{Abort}) that allows the LLM to stop when it determines the task cannot be solved.


\paragraph{Tool Review:}We conducted a multi-stage review to ensure correctness of tools. First, we programmatically verified consistency between tool and parameter names in the descriptions and the generated python code. Second, we used Gemini to automatically flag tools that might expose row-level or (individual data) and manually audited all flagged cases. We removed several functions: \texttt{sample\_rows} (non-deterministic, returns actual data rows), \texttt{compute\_residuals} (returns residuals per row), and \texttt{calculate\_row\_sum} and \texttt{calculate\_row\_mean} (return individual row-level values). After filtering, we retained 93 general-purpose tools, resulting in a final toolset of 100.

\begin{table*}[!tp]
  \centering
  \begin{tabular}{@{}p{4.5cm}rrr|rrr|r@{}}
    \toprule
    \textbf{Model} & \multicolumn{3}{c}{\textbf{With Python Code}} & \multicolumn{4}{c}{\textbf{With Tool Calls}} \\
    \cmidrule(lr){2-4} \cmidrule(lr){5-8}
    & \textbf{Easy} & \textbf{Med.} & \textbf{Hard} & \textbf{Easy} & \textbf{Med.} & \textbf{Hard} & \textbf{Hard+} \\
    \midrule
    gpt-oss-120b & 98.8 & 98.4 & 94.0 & 97.3 & 89.2 & 69.6 & 61.0 \\
    gpt-oss-20b & 95.5 & 96.2 & 90.0 & 96.1 & 84.8 & 59.7 & 53.1 \\[0.75em]

    Qwen3-Next-80B (4 bit) & 97.5 & 94.6 & 84.9 & 93.4 & 81.8 & 67.5 & 63.6 \\
    Qwen3-30B-A3B & 95.5 & 92.7 & 81.6 & 92.4 & 81.3 & 50.7 & 38.4 \\
    Qwen3-14B (awq) & 92.8 & 92.4 & 82.5 & 91.5 & 79.4 & 46.5 & 36.6 \\
    Qwen3-8B (awq) & 95.2 & 90.5 & 81.7 & 69.8 & 72.4 & 46.7 & 35.5 \\
    Qwen3-4B (awq) & 91.9 & 89.4 & 72.9 & 88.4 & 62.3 & 28.2 & 21.2 \\
    Qwen3-1.7B & 86.4 & 72.6 & 46.9 & 65.5 & 17.1 & 7.5 & 8.2 \\
    Qwen3-0.6B & 73.3 & 38.8 & 8.7 & 6.6 & 3.0 & 2.8 & 2.4 \\[0.75em]

    Llama-xLAM-2-8b-fc-r & 55.6 & 64.5 & 42.1 & 69.4 & 22.2 & 9.8 & 5.5 \\
    xLAM-2-3b-fc-r & 89.5 & 68.0 & 35.6 & 45.5 & 4.9 & 1.6 & 0.9 \\
    xLAM-2-1b-fc-r & 76.7 & 42.8 & 13.4 & 11.0 & 0.0 & 0.0 & 0.1 \\[0.75em]

    ToolACE-8B & 90.9 & 78.0 & 56.3 & 40.9 & 3.3 & 2.9 & 0.4 \\
    Hermes-2-Pro-Mistral-7B & 82.4 & 53.7 & 28.7 & 0.0 & 0.8 & 0.7 & 0.4 \\
    AI21-Jamba-Reasoning-3B & 59.1 & 53.9 & 23.4 & 11.2 & 2.2 & 0.8 & 0.7 \\
    \bottomrule
  \end{tabular}
  \caption{\textbf{Benchmarking Performance of Various LLMs on \dataname\ dataset:}  While LLMs demonstrate proficiency in code generation and simple tool usage, they lack compositional, multi-step tool-based reasoning ability, required for harder questions. Hard+: Hard questions with minor surface perturbations.  }
  \label{tab:main_results}
\end{table*}

\subsection{Multi-step Reasoning Question Generation}\label{sec:qgen}

For each tabular dataset selected in Section~\ref{sec:tabular_selection}, we use Gemini 2.5 Pro to generate multi-step reasoning questions at three difficulty levels: Easy, Medium, and Hard. We provide the model with four types of context: (1) column descriptions, (2) five randomly sampled rows, (3) definitions of the finalized set of 100 tools, and (4) one example question at the target difficulty level. We then instruct it to generate data analysis questions that meet the following requirements: (1) each question must reference the exact dataset filename and use column names or values exactly as they appear in the samples, with all such names enclosed in single quotes; (2) the final answer must be a single numerical value or string; (3) the question should be phrased naturally, as if posed by a data analyst; and (4) it must be solvable using only the provided tools. To guide the model toward questions solvable using just tools, we ask it to produce both the question and a complete solution trace showing the sequence of tool calls. These traces are not used for evaluation only as guidance to the model.

We further instruct the model to generate Easy questions solvable with up to three tool calls, Medium questions requiring at least four, and Hard questions requiring at least ten. For each dataset, the model generates five Easy, five Medium, and twenty Hard questions, each including (1) the assigned difficulty level, (2) the question, and (3) the corresponding solution trace. We use the model-assigned difficulty label as the final category. In total, this process yields 587 Easy, 569 Medium, and 2,260 Hard questions. See Table~\ref{tab:data_stats} for dataset statistics and Table~\ref{tab:indata-example} for Easy and Hard examples.

\subsection{Ground Truth Answer Generation} \label{sec:gt}

For each question generated in Section~\ref{sec:qgen}, we provide five large language models—Gemini 2.5 Pro, Gemini 2.5 Flash, Gemini 2.5 Flash Lite, GPT-5, and GPT-5 Mini—with the question and five sample rows from the corresponding CSV file. We instruct each model to generate a Python function that takes the CSV filename as its only argument and returns a single numerical value or string as the answer. We execute each generated function using the associated CSV file. If a function fails to compile or raises a runtime error, we record its output as \texttt{None}. We then filter the questions using two criteria: (1) at least four of the five implementations must produce valid (non-\texttt{None}) outputs, and (2) all valid outputs must be identical. Questions that fail either condition are discarded. We apply this strict filtering to ensure that the ground-truth answers are highly reliable. After applying these criteria, the final dataset contains 484 Easy, 369 Medium, and 1,210 Hard questions.

\section{Experiments}

We benchmark the multi-step tool-calling capabilities of 15 open-source LLMs on the \dataname\ dataset. Models are evaluated on their ability to solve data analysis tasks using just the 100 predefined tools described in Section 3.2, where they must compose sequential tool calls to arrive at final answers. As a reference comparison, we also measure performance when models generate Python code directly—an approach that provides an upper-bound baseline but introduces the security risks our tool-calling methodology is designed to eliminate.

\subsection{Models Benchmarked}

We benchmarked the following 15 diverse open-source LLMs: 

\begin{itemize}
\item \textbf{Large models:} \texttt{gpt-oss-120B}, \texttt{Qwen3-Next-80B (4-bit)}, \texttt{Qwen3-30B-A3B} — state-of-the-art Mixture-of-Expert general-purpose models exceeding 30B overall parameters~\cite{openai2025gptoss120bgptoss20bmodel,qwen3}.
\item \textbf{Medium-sized models:} \texttt{gpt-oss-20B}, \texttt{Qwen3-14B-AWQ}, \texttt{Qwen3-8B-AWQ}, \texttt{Qwen3-4B-AWQ} — general-purpose models ranging from 4B to 20B parameters.
\item \textbf{Small models:} \texttt{Qwen3-1.7B}, \texttt{Qwen3-0.6B} — lightweight models under 2B parameters.
\item \textbf{Tool-calling specialized models:} \texttt{xLAM-2} series (1B, 3B, 8B-Llama), \texttt{ToolACE-8B}, \texttt{Hermes-2-Pro-Mistral-7B} -- models finetuned for tool and API usage through specialized training datasets~\cite{liu2025toolace,zhang-etal-2025-xlam,hermes2pro_mistral7b}.
\end{itemize}

While our tool-calling approach allows proprietary models to be used on sensitive data—since raw data is never sent to external APIs—we focus on open-source models for several reasons. First, organizations working with confidential data often require on-premises deployment to avoid any potential leakage, including through aggregate statistics such as averages or counts. Second, open-source models ensure reproducibility and broader accessibility for benchmarking. All models are served using the \textsc{vLLM} infrastructure.

\subsection{Experiment Setup}\label{sec:experiment_setup}

\paragraph{Initialization:} For each question, we create a temporary directory with the relevant CSV file and initialize the conversation history with a user message containing the question and a system message instructing the model to solve it using only the available tools, one call at a time.

\paragraph{Agentic Loop:} We start the agentic loop by calling the LLM with the initialized conversation history and descriptions of all 100 tools (via the tool parameter). Once the LLM responds, we use the vLLM tool parser to extract any tool calls. If the output contains exactly one tool call, we execute the corresponding tool with the provided arguments. On success, we update the conversation history with the LLM output and tool result; on failure, we update it with the LLM output and the truncated runtime error message. If the output contains zero or multiple tool calls, we update the conversation history with the LLM output and a user message stating, “Your response did not contain exactly one tool call. Please try again.” This concludes the first turn (i.e., one an LLM call and user/tool response).

\paragraph{Termination:} The agentic loop continues until one of the following conditions is met: (1) the LLM calls a special reporting or abort tool (\texttt{report\_number}, \texttt{report\_string}, or \texttt{abort\_task}); (2) the turn count reaches the difficulty-dependent limit (20 for Easy, 30 for Medium, or 40 for Hard); or (3) the conversation history exceeds the model’s maximum context length of 32,768 tokens (\texttt{Qwen3} maximum limit). Upon termination, we clean up the temporary directory.

\paragraph{Evaluation:}After termination, we extract the predicted answer from the argument of the reporting tool (\texttt{report\_number} or \texttt{report\_string}) if invoked, or from the last tool response. A prediction is correct if it matches the ground truth—numerical answers must agree within one decimal place after rounding, and string answers must match exactly; otherwise, it is marked incorrect.

\paragraph{Python Code Generation Baseline:}For reference, we also evaluate each model’s ability to answer questions by directly generating and executing Python code. In this baseline, we prompt the LLM with the question and five sample rows from the CSV data and instruct it to generate a Python function that takes the CSV file path as its only argument and returns the final answer. We then extract and execute the code on the relevant CSV file, and treat the function’s return value as the predicted answer. If any error occurs, the predicted answer is set to \texttt{None}.

\subsection{Benchmarking Results}

Table \ref{tab:main_results} benchmarks LLM performance on \dataname. In the code-generation setting, performance is consistently high across difficulty levels (gpt-oss-120b: 98.8\% Easy, 94.0\% Hard). This strong baseline is expected given current LLMs' proficiency in code generation, and our dataset design reinforces this advantage—filtering (Section \ref{sec:gt}) ensured all questions were solvable via code, creating conditions favorable to programmatic solutions. Large models also excel on Easy questions in the tool-calling setting (gpt-oss-120b: 97.3\%), which involve simple tool selection and invocation—capabilities that current LLMs are well-optimized for and that constitute the primary focus of prior tool-use benchmarks.

However, performance on Hard questions reveals a critical capability gap. gpt-oss-120b's accuracy drops from 97.3\% (Easy) to 69.6\% (Hard), showing that while models handle individual tool operations effectively, composing them into long sequential chains for complex reasoning remains challenging. This compositional capability is also highly scale-dependent: smaller models show severe degradation, with Qwen3-1.7B declining from 65.5\% (Easy) to just 7.5\% (Hard). Furthermore, models specifically fine-tuned for tool calling on other benchmarks, such as ToolACE-8B, perform poorly on \dataname's Hard subset (2.9\%), despite achieving strong performance on BFCL and API-Bank~\cite{liu2025toolace}. This suggests that \dataname\ evaluates a distinct, compositional reasoning capability, extending beyond the focus on tool selection and invocation performance found in prior works. \noindent \textbf{Key Findings}: While LLMs demonstrate proficiency in code generation and foundational tool usage—capabilities well-addressed by existing benchmarks—our results reveal an important gap: current LLMs lack compositional, multi-step tool reasoning ability, a distinct capability that requires targeted research and development.

\begin{table*}[!htbp]
    \centering
    \label{tab:tool_comparison_multirow}
    \begin{tabular}{lcccc|ccc}
        \toprule
        \multirow{2}{*}{\textbf{Category}} & \multirow{2}{*}{\textbf{Model}} & \multicolumn{3}{c|}{\textbf{Sufficient Tools }} & \multicolumn{3}{c}{\textbf{All Tools }} \\
        \cmidrule(lr){3-5} \cmidrule(l){6-8}
        & & \textbf{Easy} & \textbf{Medium} & \textbf{Hard} & \textbf{Easy} & \textbf{Medium} & \textbf{Hard} \\
        \midrule
        \multirow{4}{*}{\textbf{Large}} 
        & gpt-oss-120B & 92.0 & 89.3 & 79.2 & \textbf{98.3} & \textbf{92.4} & \textbf{79.4} \\
        & gpt-oss-20B & 92.5 & 82.6 & 64.1 & \textbf{97.3} & \textbf{87.6} & \textbf{68.7} \\
        & Qwen3-Next-80B & 90.1 & 80.9 & 70.4 & \textbf{94.5} & \textbf{84.6} & \textbf{77.7} \\
        & Qwen3-30B & 87.6 & 82.3 & \textbf{59.2} & \textbf{93.7} & \textbf{83.4} & 58.7 \\
        \midrule
        \multirow{4}{*}{\textbf{Small}} 
        & xLAM-3B & \textbf{56.0} & \textbf{12.1} & \textbf{4.6} & 46.1 & 5.1 & 1.9 \\
        & xLAM-1B & \textbf{42.8} & \textbf{0.3} & \textbf{0.5} & 11.1 & 0.0 & 0.0 \\
        & Jamba-3B & \textbf{24.3} & \textbf{3.7} & \textbf{2.8} & 11.3 & 2.2 & 0.9 \\
        & Hermes-7B & \textbf{43.0} & \textbf{17.7} & \textbf{6.2} & 0.0 & 0.8 & 0.8 \\
        \bottomrule
    \end{tabular}
\caption{\textbf{All vs. Sufficient Tools:} Larger capable models perform better with all tools, while smaller or older models benefit from using only the sufficient ones. The table reports the percentage of correct answers on the subset containing at least one tool only solution (from Table~\ref{tab:with_hints}), therefore sufficient tools.}\label{tab:suff_tools}\end{table*}

\begin{table}[!htbp]
  \centering
  \begin{tabular}{lrrr}
    \toprule
    \textbf{Model} & \textbf{Easy} & \textbf{Medium} & \textbf{Hard}\\
    \midrule
        gpt-oss-120b & 98.3 & 94.3 & 74.9  \\
        Qwen3-Next-80B & 94.8 & 87.0 & 73.4 \\
    \midrule
        (Either One) & 98.6  & 96.7 & 85.1\\ 
    \bottomrule
  \end{tabular}
  \caption{\textbf{Performance with Hints:} LLMs performance better when code and sample table are given as Hints. (Either One) = atleast one of the two model leads to correct answer; used in \S~\ref{sec:suff_tools}.}
  \label{tab:with_hints}
\end{table}

\section{Analysis}

\subsection{Hard+: Robustness to Syntactical Variation} \label{sec:hardplus}

To evaluate robustness to small variations in questions, we construct \emph{Hard+} from the original \emph{Hard} set by introducing minor surface-level perturbations. During question generation (Section~\ref{sec:qgen}), the LLM is instructed to use column names exactly as they appear in the CSV files. In \textit{Hard+}, we modify these references as follows: (1) replace underscores or hyphens with spaces (e.g., \texttt{patient\_age} $\rightarrow$ \texttt{patient age}); (2) replace parentheses with commas (e.g., \texttt{year (2000)} $\rightarrow$ \texttt{year, 2000}); and (3) alter capitalization (e.g., \texttt{Age} $\rightarrow$ \texttt{age}). These minor edits can trigger \emph{column not found} errors. Ideally, an LLM should recover by invoking \texttt{get\_columns} to inspect the schema before retrying, requiring only one or two additional tool calls. We repeat the experiment using the exact same prompt and settings as in the original \textit{Hard} evaluation. As we can see from Table~\ref{tab:main_results}, performance drops notably—for instance, \texttt{gpt-oss-120b} accuracy decreases from 69.6\% on \textit{Hard} to 61.0\% on \textit{Hard+}. This reinforces the trend observed from \textit{Easy} to \textit{Hard}: current LLMs struggle with longer multi-step tool use, and even a few additional tool calls can significantly reduce performance.

\subsection{Improved Performance with Hints}
\label{sec:with_hints}

We test whether providing additional hints can help models improve performance. For each question, we include two hints: (i) the correct Python code solution and (ii) five sampled rows from the relevant CSV table. The model must still answer the question using tool calls, but these hints serve as structured guidance. As shown in Table~\ref{tab:with_hints}, providing hints yields consistent gains across all difficulty levels—for instance, \texttt{gpt-oss-120b} improves from 69.6\% on \textit{Hard} (without hints) to 74.9\% with hints. This result suggests a promising direction for enhancing LLMs’ multi-step tool use: rather than executing generated code, one can generate the Python solution first and provide it as hint, retaining the security benefits of controlled tool execution while improving success rates.


\subsection{Lower Bound on Completeness}
\label{sec:completeness}

During dataset construction, we retained only questions solvable by at least four of five LLMs using Python-based solutions. However, it remains uncertain whether all of these can also be solved using only the provided tools. To estimate this, we refer to the final row in Table~\ref{tab:with_hints}, labeled \emph{(Either One)}, which reports the percentage of questions correctly answered by at least one of the two models (\texttt{gpt-oss-120b} or \texttt{Qwen3-Next-80B}). This serves as a conservative \emph{lower bound} on dataset completeness relative to the current tool set—98.6\% for \textit{Easy}, 96.7\% for \textit{Medium}, and 85.1\% for \textit{Hard}. The remaining questions may still be solvable using the same tools but were not successfully solved by the current models, suggesting that completeness could increase with stronger or more diverse LLMs.

\subsection{ Performance with Sufficient Tools}\label{sec:suff_tools}

In earlier experiments, each LLM had access to all 100 tools and was expected to identify and use the relevant ones. This setup adds complexity in two ways: (1) the model must decide which tools are needed, and (2) the large tool list increases context length. Here, we analyze performance when the model is given only the \emph{sufficient subset} of tools required for each question. We first select questions with at least one known tool-only solution (98.6\%, 96.7\%, and 85.1\% for Easy, Medium, and Hard; from Table~\ref{tab:with_hints}). For each selected question, we extract the exact tools used in its verified solution and treat them as the sufficient tool set. We then re-run the agentic loop on these questions, providing the LLM with only the corresponding sufficient tools instead of the full set of 100.

\begin{figure}[!tbp]
    \centering
    \includegraphics[width=0.99\linewidth]{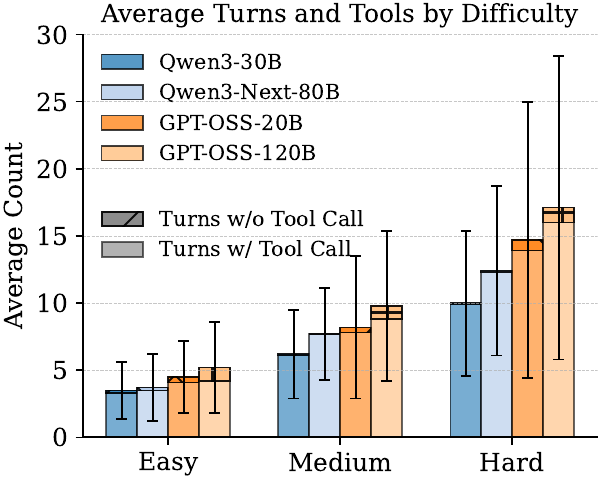}
    \caption{Average number of turns per question with and without tool calls.}
    \label{fig:tool_counts}
\end{figure}

As shown in Table~\ref{tab:suff_tools}, large models perform better with all tools, suggesting they can handle longer contexts and benefit from greater flexibility. In contrast, smaller models perform substantially better when restricted to their sufficient tool sets—for example, \texttt{xLAM-1B} improves from 11.1\% to 42.8\% on \textit{Easy}. This shows that smaller models are more sensitive to context length and benefit from shorter inputs and simpler tool selection. \textbf{Takeaway:} Large models effectively manage long-context reasoning and tool selection, benefiting from flexibility, while smaller models gain from reduced context length and constrained tool sets.

\subsection{Solution Complexity}

In Figure~\ref{fig:tool_counts}, we show, for each difficulty level and the four best-performing models, the average number of LLM turns per question, both including and excluding tool calls. We observe that (1) most turns involve tool calls, as intended, and (2) the average number of turns and tool calls increases sharply from \textit{Easy} to \textit{Hard} questions. For \textit{Hard} examples, models require on average more than 17 turns (about 16 tool calls) and often exceed 25 turns. These results confirm that \dataname\ includes complex, multi-turn questions.

\begin{table}[!tbp]
    \centering
    \label{tab:tool_performance_avg_abbr}
    \begin{tabular}{l cccc}
        \toprule
        \textbf{Model} & \multicolumn{1}{p{0.7cm}}{\centering \textbf{Corr}} & \multicolumn{1}{p{0.7cm}}{\centering \textbf{RT}} & \multicolumn{1}{p{0.7cm}}{\centering \textbf{Miss}} & \multicolumn{1}{p{0.7cm}}{\centering \textbf{Multi}} \\
        \midrule
        gpt-oss-120b & 14.6 & 1.4 & 1.04 & 0.01 \\
        gpt-oss-20b  & 12.3 & 1.6 & 0.69 & 0.12 \\
        Qwen3-30B-A3B & 8.7 & 1.2 & 0.03 & 0.09 \\
        Qwen3-Next-80B & 11.7 & 0.6 & 0.09 & 0.00 \\
        \bottomrule
    \end{tabular}
    \caption{Average number of turns per \textit{Hard} question, categorized by outcome: correct \& successful tool execution (Corr), runtime error (RT), missing tool calls (Miss), and multiple  tool calls (Multi).} \label{tab:intermediate_ea}
\end{table}

\subsection{Intermediate Error Analysis}

In Table~\ref{tab:intermediate_ea}, we report the average number of turns per \textit{Hard} question across four categories: (1) correct and successful tool executions, (2) executions that invoked an non-existent hallucinated tool or produced a runtime error, (3) responses with no tool calls, and (4) responses containing multiple tool calls. The results show that cases with missing or multiple tool calls are rare, and most tool executions complete successfully.

\begin{table}[!tbp]
    \centering
    \label{tab:tool_metrics_v3}
    \begin{tabular}{l cccc}
        \toprule
        \textbf{Model} & \multicolumn{1}{p{0.9cm}}{\centering Report Tool} & \multicolumn{1}{p{0.9cm}}{\centering Abort Tool} & \multicolumn{1}{p{0.8cm}}{\centering Max Turn} & \multicolumn{1}{p{1.0cm}}{\centering Max Length} \\
        \midrule
        gpt-120b & 76.0 & 9.0 & 15.0 & 0.0 \\
        gpt-20b  & 71.7 & 18.3 & 10.0 & 0.0 \\
        Qwen3-80B & 85.6 & 11.8 & 1.6 & 1.0 \\
        Qwen3-30B& 71.0 & 27.4 & 1.5 & 0.2 \\
        \bottomrule
    \end{tabular}
    \caption{Percentage of \textit{Hard} questions terminated under each of the 4 termination criterion}\label{tab:termination}
\end{table}

\subsection{Termination Analysis}

In Section~\ref{sec:experiment_setup}, we outlined four criteria for terminating the agentic loop. Table~\ref{tab:termination} presents, for the four best-performing models on the \textit{Hard} subset, the percentage of examples ending under each criterion. The results show that \texttt{gpt}-series models more frequently hit the maximum-turn limit, while \texttt{Qwen} models invoke the \texttt{abort\_task} condition considerably more often.

\section{Conclusion}

We introduced \dataname, a dataset for evaluating how LLMs perform on complex, multi-step data analysis tasks when restricted from directly accessing data or generating and executing code. Instead, models must solve problems exclusively through a predefined set of secure, vetted tools. We benchmarked 15 diverse LLMs and found that while current models demonstrate strong capabilities in code generation and basic tool use, they still lack robust compositional reasoning across multiple tool calls—a distinct and underexplored capability that warrants further research. We also presented several analyses highlighting key insights and directions for improving LLMs’ tool-calling abilities.

Looking ahead, an important next step is to develop methods that enhance this capability. A promising direction is introducing a planning stage before tool execution—where the model first formulates an explicit reasoning plan (e.g., in Python) and then uses it as a structured hint during inference. It is also worth exploring whether supervised fine-tuning, preference-based optimization, or inference-time techniques such as search or self-consistency can improve multi-step reasoning. While this work establishes a benchmark for current performance, future research should focus on strengthening LLMs’ ability to plan, reason, and execute through tools effectively.

\section{Ethics Statement and Limitations}

\textbf{Limitations}: Our tools are generated with the help of advanced LLMs rather than being manually implemented and verified for safety. While this approach efficiently supports quick large-scale benchmark construction, it introduces potential risks related to correctness and security that would ideally be eliminated through human-authored tools. The framework’s design—restricting data analysis tasks to predefined tools—limits the expressive potential of LLMs for solving more complex problems. Nevertheless, the executed code remains secure and isolated, preventing data corruption or leakage, even though LLMs may still produce logical or reasoning errors. Despite a stringent threshold for establishing ground truth, all reference answers ultimately originate from LLM outputs, which may introduce bias. Particularly difficult questions with inconsistent model solutions are excluded, potentially underrepresenting some challenging cases. Finally, even aggregate results can reveal sensitive information if queried extensively. Limiting the number of tool calls and truncating all tool outputs (including error messages) reduces this risk but does not eliminate it. Therefore, open-source, locally deployed LLMs are preferable for highly confidential data. Encrypting tool responses could further mitigate leakage but would add significant complexity.\\

\noindent\textbf{Ethics Statement}: To the best of our knowledge, the datasets used in this work do not contain sensitive information. While the proposed tool-based data analysis framework is not designed for any specific application, it could be applied in sensitive domains such as healthcare. We strongly encourage any future work building on our data or methods to conduct thorough quality assurance and robustness testing before deployment. All datasets and code required to reproduce our experiments will be made publicly available.

\section{Bibliographical References}\label{sec:reference}

\bibliographystyle{lrec2026-natbib}
\bibliography{lrec2026-example}

@InProceedings{DS-Agent,
  title = 	 {{DS}-Agent: Automated Data Science by Empowering Large Language Models with Case-Based Reasoning},
  author =       {Guo, Siyuan and Deng, Cheng and Wen, Ying and Chen, Hechang and Chang, Yi and Wang, Jun},
  booktitle = 	 {Proceedings of the 41st International Conference on Machine Learning},
  pages = 	 {16813--16848},
  year = 	 {2024},
  volume = 	 {235},
  series = 	 {Proceedings of Machine Learning Research},
  publisher =    {PMLR}
}

@article{qi2024cleanagent,
  title={CleanAgent: Automating Data Standardization with LLM-based Agents},
  author={Qi, Danrui and Wang, Jiannan},
  journal={arXiv preprint arXiv:2403.08291},
  year={2024}
}

@misc{zhang2025datascibenchllmagentbenchmark,
      title={DataSciBench: An LLM Agent Benchmark for Data Science}, 
      author={Dan Zhang and Sining Zhoubian and Min Cai and Fengzu Li and Lekang Yang and Wei Wang and Tianjiao Dong and Ziniu Hu and Jie Tang and Yisong Yue},
      year={2025},
      eprint={2502.13897},
      archivePrefix={arXiv},
      primaryClass={cs.CL},
      url={https://arxiv.org/abs/2502.13897}, 
}

@inproceedings{zhang-etal-2024-benchmarking-data,
    title = "Benchmarking Data Science Agents",
    author = "Zhang, Yuge  and
      Jiang, Qiyang  and
      XingyuHan, XingyuHan  and
      Chen, Nan  and
      Yang, Yuqing  and
      Ren, Kan",
    editor = "Ku, Lun-Wei  and
      Martins, Andre  and
      Srikumar, Vivek",
    booktitle = "Proceedings of the 62nd Annual Meeting of the Association for Computational Linguistics (Volume 1: Long Papers)",
    month = aug,
    year = "2024",
    address = "Bangkok, Thailand",
    publisher = "Association for Computational Linguistics",
    url = "https://aclanthology.org/2024.acl-long.308/",
    doi = "10.18653/v1/2024.acl-long.308",
    pages = "5677--5700"
}

@misc{rasheed2024largelanguagemodelsserve,
      title={Can Large Language Models Serve as Data Analysts? A Multi-Agent Assisted Approach for Qualitative Data Analysis}, 
      author={Zeeshan Rasheed and Muhammad Waseem and Aakash Ahmad and Kai-Kristian Kemell and Wang Xiaofeng and Anh Nguyen Duc and Pekka Abrahamsson},
      year={2024},
      eprint={2402.01386},
      archivePrefix={arXiv},
      primaryClass={cs.SE},
      url={https://arxiv.org/abs/2402.01386}, 
}

@misc{
li2025benchmarking,
title={Benchmarking Intelligent {LLM} Agents for Conversational Data Analysis},
author={Jinyang Li and Nan Huo and Yan Gao and Jiayi Shi and Yingxiu Zhao and Ge Qu and Bowen Qin and Xiaodong Li and Chenhao Ma and Jian-Guang Lou and Reynold Cheng},
year={2025},
url={https://openreview.net/forum?id=1zgil8py5o}
}

@misc{hong2024datainterpreterllmagent,
      title={Data Interpreter: An LLM Agent For Data Science}, 
      author={Sirui Hong and Yizhang Lin and Bang Liu and Bangbang Liu and Binhao Wu and Ceyao Zhang and Chenxing Wei and Danyang Li and Jiaqi Chen and Jiayi Zhang and Jinlin Wang and Li Zhang and Lingyao Zhang and Min Yang and Mingchen Zhuge and Taicheng Guo and Tuo Zhou and Wei Tao and Xiangru Tang and Xiangtao Lu and Xiawu Zheng and Xinbing Liang and Yaying Fei and Yuheng Cheng and Zhibin Gou and Zongze Xu and Chenglin Wu},
      year={2024},
      eprint={2402.18679},
      archivePrefix={arXiv},
      primaryClass={cs.AI},
      url={https://arxiv.org/abs/2402.18679}, 
}

@InProceedings{pmlr-v235-hu24s,
  title = 	 {{I}nfi{A}gent-{DAB}ench: Evaluating Agents on Data Analysis Tasks},
  author =       {Hu, Xueyu and Zhao, Ziyu and Wei, Shuang and Chai, Ziwei and Ma, Qianli and Wang, Guoyin and Wang, Xuwu and Su, Jing and Xu, Jingjing and Zhu, Ming and Cheng, Yao and Yuan, Jianbo and Li, Jiwei and Kuang, Kun and Yang, Yang and Yang, Hongxia and Wu, Fei},
  booktitle = 	 {Proceedings of the 41st International Conference on Machine Learning},
  pages = 	 {19544--19572},
  year = 	 {2024},
  editor = 	 {Salakhutdinov, Ruslan and Kolter, Zico and Heller, Katherine and Weller, Adrian and Oliver, Nuria and Scarlett, Jonathan and Berkenkamp, Felix},
  volume = 	 {235},
  series = 	 {Proceedings of Machine Learning Research},
  month = 	 {21--27 Jul},
  publisher =    {PMLR},
  url = 	 {https://proceedings.mlr.press/v235/hu24s.html}
}

@article{Lai2022DS1000,
  title={DS-1000: A Natural and Reliable Benchmark for Data Science Code Generation},
  author={Lai, Yuhang and Li, Chengxi and Wang, Yiming and Zhang, Tianyi and Zhong, Ruiqi and Zettlemoyer, Luke and Yih, Wen-Tau and Fried, Daniel and Wang, Sida and Yu, Tao},
  journal={ArXiv},
  year={2022},
  volume={abs/2211.11501}
}

@article{chen2021codex,
  title={Evaluating Large Language Models Trained on Code},
  author={Mark Chen and Jerry Tworek and Heewoo Jun and Qiming Yuan and Henrique Pond{\'e} and Jared Kaplan and Harrison Edwards and Yura Burda and Nicholas Joseph and Greg Brockman and Alex Ray and Raul Puri and Gretchen Krueger and Michael Petrov and Heidy Khlaaf and Girish Sastry and Pamela Mishkin and Brooke Chan and Scott Gray and Nick Ryder and Mikhail Pavlov and Alethea Power and Lukasz Kaiser and Mo Bavarian and Clemens Winter and Phil Tillet and Felipe Petroski Such and David W. Cummings and Matthias Plappert and Fotios Chantzis and Elizabeth Barnes and Ariel Herbert-Voss and William H. Guss and Alex Nichol and Igor Babuschkin and Suchir Balaji and Shantanu Jain and Andrew Carr and Jan Leike and Josh Achiam and Vedant Misra and Evan Morikawa and Alec Radford and Matthew M. Knight and Miles Brundage and Mira Murati and Katie Mayer and Peter Welinder and Bob McGrew and Dario Amodei and Sam McCandlish and Ilya Sutskever and Wojciech Zaremba},
  journal={ArXiv},
  year={2021},
  volume={abs/2107.03374},
  url={https://api.semanticscholar.org/CorpusID:235755472}
}

@misc{austin2021programsynthesislargelanguage,
      title={Program Synthesis with Large Language Models}, 
      author={Jacob Austin and Augustus Odena and Maxwell Nye and Maarten Bosma and Henryk Michalewski and David Dohan and Ellen Jiang and Carrie Cai and Michael Terry and Quoc Le and Charles Sutton},
      year={2021},
      eprint={2108.07732},
      archivePrefix={arXiv},
      primaryClass={cs.PL},
      url={https://arxiv.org/abs/2108.07732}, 
}

@misc{li2023llmservedatabaseinterface,
      title={Can LLM Already Serve as A Database Interface? A BIg Bench for Large-Scale Database Grounded Text-to-SQLs}, 
      author={Jinyang Li and Binyuan Hui and Ge Qu and Jiaxi Yang and Binhua Li and Bowen Li and Bailin Wang and Bowen Qin and Rongyu Cao and Ruiying Geng and Nan Huo and Xuanhe Zhou and Chenhao Ma and Guoliang Li and Kevin C. C. Chang and Fei Huang and Reynold Cheng and Yongbin Li},
      year={2023},
      eprint={2305.03111},
      archivePrefix={arXiv},
      primaryClass={cs.CL},
      url={https://arxiv.org/abs/2305.03111}, 
}

@inproceedings{Yu&al.18c,
  title     = {Spider: A Large-Scale Human-Labeled Dataset for Complex and Cross-Domain Semantic Parsing and Text-to-SQL Task},
  author    = {Tao Yu and Rui Zhang and Kai Yang and Michihiro Yasunaga and Dongxu Wang and Zifan Li and James Ma and Irene Li and Qingning Yao and Shanelle Roman and Zilin Zhang and Dragomir Radev},
  booktitle = "Proceedings of the 2018 Conference on Empirical Methods in Natural Language Processing",
  address   = "Brussels, Belgium",
  publisher = "Association for Computational Linguistics",
  year      = 2018
}

@article{hendrycksapps2021,
  title={Measuring Coding Challenge Competence With APPS},
  author={Dan Hendrycks and Steven Basart and Saurav Kadavath and Mantas Mazeika and Akul Arora and Ethan Guo and Collin Burns and Samir Puranik and Horace He and Dawn Song and Jacob Steinhardt},
  journal={NeurIPS},
  year={2021}
}

@article{DBLP:journals/corr/abs-2102-04664,
  author    = {Shuai Lu and
               Daya Guo and
               Shuo Ren and
               Junjie Huang and
               Alexey Svyatkovskiy and
               Ambrosio Blanco and
               Colin B. Clement and
               Dawn Drain and
               Daxin Jiang and
               Duyu Tang and
               Ge Li and
               Lidong Zhou and
               Linjun Shou and
               Long Zhou and
               Michele Tufano and
               Ming Gong and
               Ming Zhou and
               Nan Duan and
               Neel Sundaresan and
               Shao Kun Deng and
               Shengyu Fu and
               Shujie Liu},
  title     = {CodeXGLUE: {A} Machine Learning Benchmark Dataset for Code Understanding
               and Generation},
  journal   = {CoRR},
  volume    = {abs/2102.04664},
  year      = {2021}
}

@InProceedings{Yu&al.19,
  title     = {SParC: Cross-Domain Semantic Parsing in Context},
  author    = {Tao Yu and Rui Zhang and Michihiro Yasunaga and Yi Chern Tan and Xi Victoria Lin and Suyi Li and Heyang Er, Irene Li and Bo Pang and Tao Chen and Emily Ji and Shreya Dixit and David Proctor and Sungrok Shim and Jonathan Kraft, Vincent Zhang and Caiming Xiong and Richard Socher and Dragomir Radev},
  booktitle = {Proceedings of the 57th Annual Meeting of the Association for Computational Linguistics},
  year      = {2019},
  address   = {Florence, Italy},
  publisher = {Association for Computational Linguistics}
}

@inproceedings{yu-etal-2019-cosql,
    title = "{C}o{SQL}: A Conversational Text-to-{SQL} Challenge Towards Cross-Domain Natural Language Interfaces to Databases",
    author = "Yu, Tao  and
      Zhang, Rui  and
      Er, Heyang  and
      Li, Suyi  and
      Xue, Eric  and
      Pang, Bo  and
      Lin, Xi Victoria  and
      Tan, Yi Chern  and
      Shi, Tianze  and
      Li, Zihan  and
      Jiang, Youxuan  and
      Yasunaga, Michihiro  and
      Shim, Sungrok  and
      Chen, Tao  and
      Fabbri, Alexander  and
      Li, Zifan  and
      Chen, Luyao  and
      Zhang, Yuwen  and
      Dixit, Shreya  and
      Zhang, Vincent  and
      Xiong, Caiming  and
      Socher, Richard  and
      Lasecki, Walter  and
      Radev, Dragomir",
    editor = "Inui, Kentaro  and
      Jiang, Jing  and
      Ng, Vincent  and
      Wan, Xiaojun",
    booktitle = "Proceedings of the 2019 Conference on Empirical Methods in Natural Language Processing and the 9th International Joint Conference on Natural Language Processing (EMNLP-IJCNLP)",
    month = nov,
    year = "2019",
    address = "Hong Kong, China",
    publisher = "Association for Computational Linguistics",
    url = "https://aclanthology.org/D19-1204/",
    doi = "10.18653/v1/D19-1204",
    pages = "1962--1979"
}

@inproceedings{yin-etal-2023-natural,
    title = "Natural Language to Code Generation in Interactive Data Science Notebooks",
    author = "Yin, Pengcheng  and
      Li, Wen-Ding  and
      Xiao, Kefan  and
      Rao, Abhishek  and
      Wen, Yeming  and
      Shi, Kensen  and
      Howland, Joshua  and
      Bailey, Paige  and
      Catasta, Michele  and
      Michalewski, Henryk  and
      Polozov, Oleksandr  and
      Sutton, Charles",
    editor = "Rogers, Anna  and
      Boyd-Graber, Jordan  and
      Okazaki, Naoaki",
    booktitle = "Proceedings of the 61st Annual Meeting of the Association for Computational Linguistics (Volume 1: Long Papers)",
    month = jul,
    year = "2023",
    address = "Toronto, Canada",
    publisher = "Association for Computational Linguistics",
    url = "https://aclanthology.org/2023.acl-long.9/",
    doi = "10.18653/v1/2023.acl-long.9",
    pages = "126--173"
}

@inproceedings{dataprepeda2021,
  author    = {Jinglin Peng and Weiyuan Wu and Brandon Lockhart and Song Bian and Jing Nathan Yan and Linghao Xu and Zhixuan Chi and Jeffrey M. Rzeszotarski and Jiannan Wang},
  title     = {DataPrep.EDA: Task-Centric Exploratory Data Analysis for Statistical Modeling in Python},
  booktitle = {Proceedings of the 2021 International Conference on Management of Data (SIGMOD '21), June 20--25, 2021, Virtual Event, China},
  year      = {2021}
}

@article{chatdev,
    title = {ChatDev: Communicative Agents for Software Development},
    author = {Chen Qian and Wei Liu and Hongzhang Liu and Nuo Chen and Yufan Dang and Jiahao Li and Cheng Yang and Weize Chen and Yusheng Su and Xin Cong and Juyuan Xu and Dahai Li and Zhiyuan Liu and Maosong Sun},
    journal = {arXiv preprint arXiv:2307.07924},
    url = {https://arxiv.org/abs/2307.07924},
    year = {2023}
}

@article{colearning,
    title = {Experiential Co-Learning of Software-Developing Agents},
    author = {Chen Qian and Yufan Dang and Jiahao Li and Wei Liu and Zihao Xie and Yifei Wang and Weize Chen and Cheng Yang and Xin Cong and Xiaoyin Che and Zhiyuan Liu and Maosong Sun},
    journal = {arXiv preprint arXiv:2312.17025},
    url = {https://arxiv.org/abs/2312.17025},
    year = {2023}
}

@article{macnet,
    title={Scaling Large-Language-Model-based Multi-Agent Collaboration},
    author={Chen Qian and Zihao Xie and Yifei Wang and Wei Liu and Yufan Dang and Zhuoyun Du and Weize Chen and Cheng Yang and Zhiyuan Liu and Maosong Sun},
    journal={arXiv preprint arXiv:2406.07155},
    url = {https://arxiv.org/abs/2406.07155},
    year={2024}
}

@article{iagents,
    title={Autonomous Agents for Collaborative Task under Information Asymmetry},
    author={Wei Liu and Chenxi Wang and Yifei Wang and Zihao Xie and Rennai Qiu and Yufan Dnag and Zhuoyun Du and Weize Chen and Cheng Yang and Chen Qian},
    journal={arXiv preprint arXiv:2406.14928},
    url = {https://arxiv.org/abs/2406.14928},
    year={2024}
}

@article{zhang2023mlcopilot,
    title={MLCopilot: Unleashing the Power of Large Language Models in Solving Machine Learning Tasks},
    author={Zhang, Lei and Zhang, Yuge and Ren, Kan and Li, Dongsheng and Yang, Yuqing},
    journal={arXiv preprint arXiv:2304.14979},
    year={2023}
}

@inproceedings{
berkeley-function-calling-leaderboard,
title={The Berkeley Function Calling Leaderboard ({BFCL}): From Tool Use to Agentic Evaluation of Large Language Models},
author={Shishir G Patil and Huanzhi Mao and Fanjia Yan and Charlie Cheng-Jie Ji and Vishnu Suresh and Ion Stoica and Joseph E. Gonzalez},
booktitle={Forty-second International Conference on Machine Learning},
year={2025},
url={https://openreview.net/forum?id=2GmDdhBdDk}
}

@misc{patil2023gorillalargelanguagemodel,
      title={Gorilla: Large Language Model Connected with Massive APIs}, 
      author={Shishir G. Patil and Tianjun Zhang and Xin Wang and Joseph E. Gonzalez},
      year={2023},
      eprint={2305.15334},
      archivePrefix={arXiv},
      primaryClass={cs.CL},
      url={https://arxiv.org/abs/2305.15334}, 
}

@misc{xu2023toolmanipulationcapabilityopensource,
      title={On the Tool Manipulation Capability of Open-source Large Language Models}, 
      author={Qiantong Xu and Fenglu Hong and Bo Li and Changran Hu and Zhengyu Chen and Jian Zhang},
      year={2023},
      eprint={2305.16504},
      archivePrefix={arXiv},
      primaryClass={cs.CL},
      url={https://arxiv.org/abs/2305.16504}, 
}

@misc{tang2023toolalpaca,
      title={ToolAlpaca: Generalized Tool Learning for Language Models with 3000 Simulated Cases}, 
      author={Qiaoyu Tang and Ziliang Deng and Hongyu Lin and Xianpei Han and Qiao Liang and Le Sun},
      year={2023},
      eprint={2306.05301},
      archivePrefix={arXiv},
      primaryClass={cs.CL}
}

@inproceedings{
qin2024toolllm,
title={Tool{LLM}: Facilitating Large Language Models to Master 16000+ Real-world {API}s},
author={Yujia Qin and Shihao Liang and Yining Ye and Kunlun Zhu and Lan Yan and Yaxi Lu and Yankai Lin and Xin Cong and Xiangru Tang and Bill Qian and Sihan Zhao and Lauren Hong and Runchu Tian and Ruobing Xie and Jie Zhou and Mark Gerstein and dahai li and Zhiyuan Liu and Maosong Sun},
booktitle={The Twelfth International Conference on Learning Representations},
year={2024},
url={https://openreview.net/forum?id=dHng2O0Jjr}
}

@misc{li2023apibankcomprehensivebenchmarktoolaugmented,
      title={API-Bank: A Comprehensive Benchmark for Tool-Augmented LLMs}, 
      author={Minghao Li and Yingxiu Zhao and Bowen Yu and Feifan Song and Hangyu Li and Haiyang Yu and Zhoujun Li and Fei Huang and Yongbin Li},
      year={2023},
      eprint={2304.08244},
      archivePrefix={arXiv},
      primaryClass={cs.CL},
      url={https://arxiv.org/abs/2304.08244}, 
}

@misc{zhuang2023toolqadatasetllmquestion,
      title={ToolQA: A Dataset for LLM Question Answering with External Tools}, 
      author={Yuchen Zhuang and Yue Yu and Kuan Wang and Haotian Sun and Chao Zhang},
      year={2023},
      eprint={2306.13304},
      archivePrefix={arXiv},
      primaryClass={cs.CL},
      url={https://arxiv.org/abs/2306.13304}, 
}

@inproceedings{
liu2025toolace,
title={Tool{ACE}: Winning the Points of {LLM} Function Calling},
author={Weiwen Liu and Xu Huang and Xingshan Zeng and xinlong hao and Shuai Yu and Dexun Li and Shuai Wang and Weinan Gan and Zhengying Liu and Yuanqing Yu and Zezhong WANG and Yuxian Wang and Wu Ning and Yutai Hou and Bin Wang and Chuhan Wu and Wang Xinzhi and Yong Liu and Yasheng Wang and Duyu Tang and Dandan Tu and Lifeng Shang and Xin Jiang and Ruiming Tang and Defu Lian and Qun Liu and Enhong Chen},
booktitle={The Thirteenth International Conference on Learning Representations},
year={2025},
url={https://openreview.net/forum?id=8EB8k6DdCU}
}

@misc{farn2023tooltalkevaluatingtoolusageconversational,
      title={ToolTalk: Evaluating Tool-Usage in a Conversational Setting}, 
      author={Nicholas Farn and Richard Shin},
      year={2023},
      eprint={2311.10775},
      archivePrefix={arXiv},
      primaryClass={cs.CL},
      url={https://arxiv.org/abs/2311.10775}, 
}

@misc{k2025studyleveragingsearchselffeedback,
      title={A Study on Leveraging Search and Self-Feedback for Agent Reasoning}, 
      author={Karthikeyan K and Michelle Yuan and Elman Mansimov and Katerina Margatina and Anurag Pratik and Daniele Bonadiman and Monica Sunkara and Yi Zhang and Yassine Benajiba},
      year={2025},
      eprint={2502.12094},
      archivePrefix={arXiv},
      primaryClass={cs.AI},
      url={https://arxiv.org/abs/2502.12094}, 
}

@article{jiang2024survey,
  title={A Survey on Large Language Models for Code Generation},
  author={Jiang, Juyong and Wang, Fan and Shen, Jiasi and Kim, Sungju and Kim, Sunghun},
  journal={arXiv preprint arXiv:2406.00515},
  year={2024}
}

@article{le2022coderl,
  title={Coderl: Mastering code generation through pretrained models and deep reinforcement learning},
  author={Le, Hung and Wang, Yue and Gotmare, Akhilesh Deepak and Savarese, Silvio and Hoi, Steven Chu Hong},
  journal={Advances in Neural Information Processing Systems},
  volume={35},
  pages={21314--21328},
  year={2022}
}

@inproceedings{ahn-etal-2024-large,
    title = "Large Language Models for Mathematical Reasoning: Progresses and Challenges",
    author = "Ahn, Janice  and
      Verma, Rishu  and
      Lou, Renze  and
      Liu, Di  and
      Zhang, Rui  and
      Yin, Wenpeng",
    editor = "Falk, Neele  and
      Papi, Sara  and
      Zhang, Mike",
    booktitle = "Proceedings of the 18th Conference of the European Chapter of the Association for Computational Linguistics: Student Research Workshop",
    month = mar,
    year = "2024",
    address = "St. Julian{'}s, Malta",
    publisher = "Association for Computational Linguistics",
    url = "https://aclanthology.org/2024.eacl-srw.17/",
    pages = "225--237"
}

@inproceedings{pan-etal-2023-logic,
    title = "Logic-{LM}: Empowering Large Language Models with Symbolic Solvers for Faithful Logical Reasoning",
    author = "Pan, Liangming  and
      Albalak, Alon  and
      Wang, Xinyi  and
      Wang, William",
    editor = "Bouamor, Houda  and
      Pino, Juan  and
      Bali, Kalika",
    booktitle = "Findings of the Association for Computational Linguistics: EMNLP 2023",
    month = dec,
    year = "2023",
    address = "Singapore",
    publisher = "Association for Computational Linguistics",
    url = "https://aclanthology.org/2023.findings-emnlp.248/",
    doi = "10.18653/v1/2023.findings-emnlp.248",
    pages = "3806--3824"
}

@article{yang2024matplotagent,
  title={Matplotagent: Method and evaluation for llm-based agentic scientific data visualization},
  author={Yang, Zhiyu and Zhou, Zihan and Wang, Shuo and Cong, Xin and Han, Xu and Yan, Yukun and Liu, Zhenghao and Tan, Zhixing and Liu, Pengyuan and Yu, Dong and others},
  journal={arXiv preprint arXiv:2402.11453},
  year={2024}
}

@article{cascella2023evaluating,
  title={Evaluating the feasibility of ChatGPT in healthcare: an analysis of multiple clinical and research scenarios},
  author={Cascella, Marco and Montomoli, Jonathan and Bellini, Valentina and Bignami, Elena},
  journal={Journal of medical systems},
  volume={47},
  number={1},
  pages={33},
  year={2023},
  publisher={Springer}
}

@inproceedings{li2023large,
  title={Large language models in finance: A survey},
  author={Li, Yinheng and Wang, Shaofei and Ding, Han and Chen, Hang},
  booktitle={Proceedings of the fourth ACM international conference on AI in finance},
  pages={374--382},
  year={2023}
}

@article{yao2024survey,
  title={A survey on large language model (llm) security and privacy: The good, the bad, and the ugly},
  author={Yao, Yifan and Duan, Jinhao and Xu, Kaidi and Cai, Yuanfang and Sun, Zhibo and Zhang, Yue},
  journal={High-Confidence Computing},
  pages={100211},
  year={2024},
  publisher={Elsevier}
}

@article{das2025security,
  title={Security and privacy challenges of large language models: A survey},
  author={Das, Badhan Chandra and Amini, M Hadi and Wu, Yanzhao},
  journal={ACM Computing Surveys},
  volume={57},
  number={6},
  pages={1--39},
  year={2025},
  publisher={ACM New York, NY}
}

@misc{basic2025largelanguagemodelscode,
      title={Large Language Models and Code Security: A Systematic Literature Review}, 
      author={Enna Basic and Alberto Giaretta},
      year={2025},
      eprint={2412.15004},
      archivePrefix={arXiv},
      primaryClass={cs.CR},
      url={https://arxiv.org/abs/2412.15004}, 
}

@misc{pearce2022asleep,
      title={Asleep at the Keyboard? Assessing the Security of GitHub Copilot's Code Contributions}, 
      author={Hammond Pearce and Baleegh Ahmad and Benjamin Tan and Brendan Dolan-Gavitt and Ramesh Karri},
      year={2021},
      eprint={2108.09293},
      archivePrefix={arXiv},
      primaryClass={cs.CR},
      url={https://arxiv.org/abs/2108.09293}, 
}

@inproceedings{perry2023users, series={CCS ’23},
   title={Do Users Write More Insecure Code with AI Assistants?},
   url={http://dx.doi.org/10.1145/3576915.3623157},
   DOI={10.1145/3576915.3623157},
   booktitle={Proceedings of the 2023 ACM SIGSAC Conference on Computer and Communications Security},
   publisher={ACM},
   author={Perry, Neil and Srivastava, Megha and Kumar, Deepak and Boneh, Dan},
   year={2023},
   month=nov, pages={2785–2799},
   collection={CCS ’23} }

@misc{carlini2021extracting,
      title={Extracting Training Data from Large Language Models}, 
      author={Nicholas Carlini and Florian Tramer and Eric Wallace and Matthew Jagielski and Ariel Herbert-Voss and Katherine Lee and Adam Roberts and Tom Brown and Dawn Song and Ulfar Erlingsson and Alina Oprea and Colin Raffel},
      year={2021},
      eprint={2012.07805},
      archivePrefix={arXiv},
      primaryClass={cs.CR},
      url={https://arxiv.org/abs/2012.07805}, 
}

@misc{openai2025gptoss120bgptoss20bmodel,
      title={gpt-oss-120b and gpt-oss-20b Model Card}, 
      author={OpenAI},
      year={2025},
      eprint={2508.10925},
      archivePrefix={arXiv},
      primaryClass={cs.CL},
      url={https://arxiv.org/abs/2508.10925}, 
}

@article{qwen3,
    title={Qwen3 Technical Report}, 
    author={An Yang and Anfeng Li and Baosong Yang and Beichen Zhang and Binyuan Hui and Bo Zheng and Bowen Yu and Chang Gao and Chengen Huang and Chenxu Lv and Chujie Zheng and Dayiheng Liu and Fan Zhou and Fei Huang and Feng Hu and Hao Ge and Haoran Wei and Huan Lin and Jialong Tang and Jian Yang and Jianhong Tu and Jianwei Zhang and Jianxin Yang and Jiaxi Yang and Jing Zhou and Jingren Zhou and Junyang Lin and Kai Dang and Keqin Bao and Kexin Yang and Le Yu and Lianghao Deng and Mei Li and Mingfeng Xue and Mingze Li and Pei Zhang and Peng Wang and Qin Zhu and Rui Men and Ruize Gao and Shixuan Liu and Shuang Luo and Tianhao Li and Tianyi Tang and Wenbiao Yin and Xingzhang Ren and Xinyu Wang and Xinyu Zhang and Xuancheng Ren and Yang Fan and Yang Su and Yichang Zhang and Yinger Zhang and Yu Wan and Yuqiong Liu and Zekun Wang and Zeyu Cui and Zhenru Zhang and Zhipeng Zhou and Zihan Qiu},
    journal = {arXiv preprint arXiv:2505.09388},
    year={2025}
}

@inproceedings{zhang-etal-2025-xlam,
    title = "x{LAM}: A Family of Large Action Models to Empower {AI} Agent Systems",
    author = "Zhang, Jianguo  and
      Lan, Tian  and
      Zhu, Ming  and
      Liu, Zuxin  and
      Hoang, Thai Quoc  and
      Kokane, Shirley  and
      Yao, Weiran  and
      Tan, Juntao  and
      Prabhakar, Akshara  and
      Chen, Haolin  and
      Liu, Zhiwei  and
      Feng, Yihao  and
      Awalgaonkar, Tulika Manoj  and
      R N, Rithesh  and
      Chen, Zeyuan  and
      Xu, Ran  and
      Niebles, Juan Carlos  and
      Heinecke, Shelby  and
      Wang, Huan  and
      Savarese, Silvio  and
      Xiong, Caiming",
    editor = "Chiruzzo, Luis  and
      Ritter, Alan  and
      Wang, Lu",
    booktitle = "Proceedings of the 2025 Conference of the Nations of the Americas Chapter of the Association for Computational Linguistics: Human Language Technologies (Volume 1: Long Papers)",
    month = apr,
    year = "2025",
    address = "Albuquerque, New Mexico",
    publisher = "Association for Computational Linguistics",
    url = "https://aclanthology.org/2025.naacl-long.578/",
    doi = "10.18653/v1/2025.naacl-long.578",
    pages = "11583--11597",
    ISBN = "979-8-89176-189-6"
}

@misc{hermes2pro_mistral7b,
  title        = {Hermes-2-Pro-Mistral-7B},
  author       = {interstellarninja and Teknium and theemozilla and karan4d and huemin\_art},
  howpublished = {\url{https://huggingface.co/NousResearch/Hermes-2-Pro-Mistral-7B}},
  note         = {Accessed: 2025-10-25},
year={2024},
}

\section{Language Resource References}
\label{lr:ref}
\bibliographystylelanguageresource{lrec2026-natbib}
\bibliographylanguageresource{languageresource}

\end{document}